# Preparing Weather Data for Real-Time Building Energy Simulation

Maryam MeshkinKiya[1], Riccardo Paolini[2]
[1]Politecnico di Milano, Milan, Italy
[2]University of New South Wales, Sydney, Australia

## Abstract

This study introduces a framework for quality control of measured weather data, including anomaly detection, and infilling missing values. Weather data is a fundamental input to building performance simulations, in which anomalous values defect the results while missing data lead to an unexpected termination of the simulation process. Traditionally, infilling missing values in weather data is performed through periodic or linear interpolations. However, when missing values exceed many consecutive hours, the accuracy of traditional methods is subject to debate. This study demonstrates how Neural Networks can increase the accuracy of data imputation when compared to other supervised learning methods. The framework is validated by predicting missing temperature and relative humidity data for an observation site, through a network of nearby weather stations in Milan, Italy. Results show that the proposed method can facilitate real-time building simulations with accurate and rapid quality control.

## Introduction

The estimation of the energy performance of buildings is conducted with a variety of different methods and tools. While software developers try to improve the building energy calculation engines, differences between simulated and real building energy consumption are repeatedly reported (Menezes, Cripps, Bouchlaghem, & Buswell, 2012). This mismatch is known as the performance gap, and different sources are recognized as contributing factors (Bordass, Cohen, Standeven, & Leaman, 2001). Inaccuracy in the input parameters is one of the most significant factors affecting the performance gap (De Wilde, 2014; Zou, Wagle, & Alam, 2019). The inputs of building energy simulations consist of design, climatic and operation parameters (Nagpal, Mueller, Aijazi, & Reinhart, 2019). The climatic parameters are defined through a dataset with different weather parameters, typically with an hourly interval. These datasets are collected based on long-term weather measurements, the most probable months are selected, and the resulting weather files are known as Typical Meteorological Year (TMY). The accuracy of these datasets has been challenged in recent studies. Since typical years select the most frequently occurring conditions out of years of measured data, extreme climatic conditions in specific years are not considered in the datasets (Siu & Liao, 2020). Moreover, the measurements are mainly performed in rural or semi-urban areas (like airports), therefore, the Urban Heat Island effect (UHI) is neglected in these datasets (Wang, Mathew, & Pang, 2012).

New approaches in the field of building energy simulation encourage the application of unfiltered weather data series as well as long periods (Cuerda, Guerra-Santin, Sendra, & Neila, 2020; Jentsch, Bahaj, & James, 2008). Using the raw weather measurements directly from weather stations brings new challenges, since these data commonly suffer from missing values. Moreover, due to maintenance or instrumental errors, the data needs regular quality control tests to ensure more accurate weather datasets (Estévez, Gavilán, & Giráldez, 2011).

Different methods are proposed for infilling missing weather data. Empirical methods, statistical and function fitting methods have been discussed for a variety of aspects regarding filling missing values (Hasanpour Kashani & Dinpashoh, 2012). Studies have compared the performance of these methods and their application for filling climatic data. Acceptable performance of empirical methods and function fitting methods, including interpolation and neural networks, are studied in the condition of systematic missing values (Kornelsen, Coulibaly, & Asce, 2014). Multiple regression is recognized as a reliable method among statistical methods (Doreswamy, Gad, & Manjunatha, 2017). The performance of empirical methods can be affected by the temporal resolution of continuous data, and therefore, poor performances have been reported for infilling missing data within daily profiles (Shabalala, Moeletsi, Tongwane, & Mazibuko, 2019)

This paper shows how neural networks can support real-time building energy simulation to predict missing weather parameters as input parameters. Moreover, a robust comparison between different supervised machine learning methods is performed in order to highlight the benefits and drawbacks of using neural networks. This study only contrasts neural networks with other machine learning methods as their benefits over traditional infilling methods have already discussed in previous papers. Missing values of two weather parameters, i.e. temperature (T) and relative humidity (RH), are targeted here. The methodology of this study is validated for a network of weather stations in northern Italy. The assumption of this study is to present a general framework for infilling missing values; therefore, the method can be expanded to other climatic parameters and weather stations.

As an original contribution, this paper provides a framework for infilling missing climatic values in real-time, by resorting to nearby weather stations. This is particularly important for real-time building energy simulation, where anomalous and missing values can distort the results or terminate of the simulation process. The framework is validate with a real case study and contrasted against other popular infilling methods.

## Main body

### Data description

The value of hourly measured temperature, and relative humidity for nine years (from 2000 until 2008) is collected from seven different weather stations within the Milan city (MI, 45.4642° N, 9.1900° E). These measured data are extracted from the network of ARPA Lombardia ("Regional Agency for the Protection of the Environment of Lombardy," n.d.). Figure 1 shows the position of weather stations under study. In the initial analysis of the data, an irregular pattern in relative humidity is observed for 2000-2004. Accordingly, relative humidity is only analysed and tested for four years (2005-2008). The training-set of temperature consists of the whole nine years. As a primary quality control test on the original data, the anomalous and missing values are (as a standard practice) replaced with the value of "-999". This will facilitate distinguishing missing data and outliers while the entire array can still be treated as a numeric matrix.

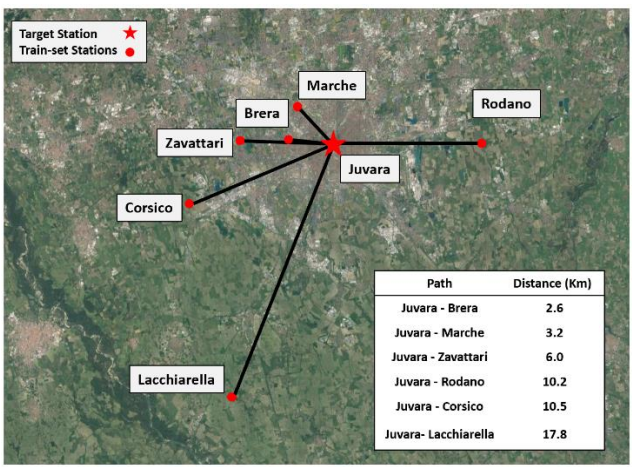

*Figure 1- The network of weather stations of this study, Milano (MI, 45.4642° N, 9.1900° E). The numbers refer to distances between the Target station and other stations in the network*

### Methods

Neural Networks (NN) – which can be categorized in the supervised function fitting groups – have been more popular in recent studies concerning climatic parameter prediction. Most of the studies in this area resort to a network of weather stations to predict missing values. This method is applied for imputation of missing temperature, solar radiation, as well as wind speed (Tardivo & Berti, 2012) (Yadav & Chandel, 2014) (Bilgili, Sahin, & Yasar, 2007).

The estimation of missing values among the datasets through machine learning methods demands a series of correct samples. Each sample consists of a set of input features and a target. The performance of machine learning methods depends on selecting correct input features, the suitable training function, as well as adequate number of samples for training and testing the network (Khayatian, Sarto, & Dall'O', 2016).

Hourly data of temperature or relative humidity mainly shape the input and target samples of this study. The input features consist of different time-steps of input samples from all weather stations in the network, except the target station. The date and hour of measurements is also included in the inputs. Finally, the target of prediction is the hourly data of a different weather station. More clarifications are provided in Table 1.

In the attempts to find the best group of input features, it is found that missing values (-999) in the input data can notably affect the performance of the network; i.e., the model is unable to ignore anomalous (-999) values as inputs. Regarding that for each input sample (hourly temperature/ relative humidity), we add a Boolean vector, where the value of "0" corresponds to anomalous value (-999), and the value "1" corresponds to non-anomalous measurements. These extra sets of Boolean features are hereafter called Logic Anomaly Indicators (LAI). Therefore, additional vectors with the same size of the main inputs are added to ensure that the weighting procedure will detect anomalies while training the neural net. Figure 2 shows the total number of input features summed up to 39.

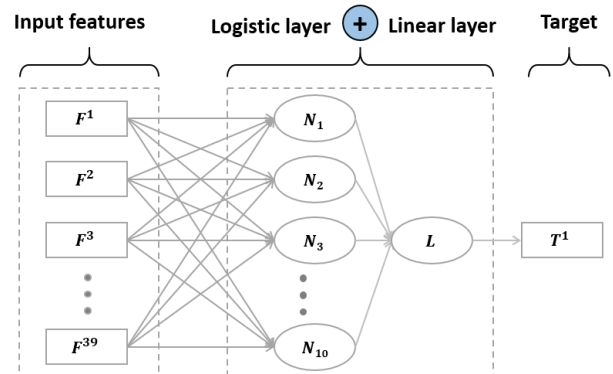

*Figure 2 - Diagram of a vanilla neural network*

The input features of studied methods are presented in Table 2. Accordingly, the input features can be categorized in three: samples of measured weather parameters (here, e.g. temperature and RH), related date of each sample, as well as Logic Anomaly Indicators (LAI). Since six weather stations are handles as inputs and based on Table 1, 18 features are related to category 1, three features are related to date (MM, DD, HH) and, another 18 features are related to LAI Boolean indicators.

*Table 1- Input feature for predicting missing temperature in target station*

| Input features | Comment |
| --- | --- |
| T-in $_{h-2}$ | Temperature in -2 hours of target hour |
| T-in $_{h-1}$ | Temperature in -1 hour of the target hour |
| T-in $_h$ | Temperature in the target hour |
| MM | |
| DD | Date of target hour |
| HH | |
| LAI $_{h-2}$ | |
| LAI $_{h-1}$ | Logic Anomaly Indicator (LAI) for $T_{h-2}$, $T_{h-1}$, $T_h$ |
| LAI $_h$ | |
| **Target** | **Comment** |
| T-out $_h$ | Temperature in the target hour |

A multilayer perceptron (MLP), which in this study is a Back-Propagated Neural Network (BPNN) is trained by using the Levenberg-Marquardt optimization algorithm. To prevent confusion during the training process, the missing values (-999) are eliminated from the target dataset. To keep the size of the input and target matrices consistent, the rows which contain (-999) in the target also removed from the input dataset.

**Other machine learning methods**

Supervised learning is a well-known approach for prediction of continuous variables (Rasmussen & Williams, 2006). Like NN which is a supervised learning algorithm, other models use a reliable set of samples including inputs and target(s) (or responses) to train a model for predicting missing values. Similarly, these methods can operate with multiple input features to train the prediction model. However, each model differs in the internal fitting algorithm. In this study, we contrast five different regression models against NN. The models are: 1) Linear regression (LR), 2) Regression Tree (RT), 3) Support Vector Regression (SVR), 4) Gaussian Process Regression (GPR), and 5) Ensemble Tree (ET).

1) Linear regression (LR) is among the popular methods to infill missing climatic parameters (Dumedah & Coulibaly, 2011; Hasanpour Kashani & Dinpashoh, 2012; Tardivo & Berti, 2014) This method, which is also known as multilinear regression, learns a linear fit on multiple inputs to predict targets, which in the case of this study refers to the missing climatic values.

2) Regression trees (RT) are also a group of machine learning methods that can predict responses based on input data. Regression trees divide the root data (inputs) into partitions and try to train a prediction model for each partition (Loh, 2011).

3) Support Vector Regressions (SVR) operate based on kernel function techniques and are considered as non-parametric models. The method is recognized as one of the most reliable methods for imputation of missing data (Aydilek & Arslan, 2013).

4) Gaussian Process Regressions (GPR) are a sub-category of bayesian machine learning methods that find the latent variables to predict the response to input data. Similar to SVR, these methods are also considered as kernel-based and non-parametric models (Rasmussen & Williams, 2006).

5) Ensemble Tree (ET) is a powerful version of decision trees that works through combining other machine learning methods such as multiple linear regression to perform a prediction (Seni & Elder, 2010).

All supervised learning methods including neural networks are implemented in MATLAB R2019b.

In order to investigate on the performance of each method, two strategies are adopted. First, separating the test-set: for observing the accuracy of trained models, three notable periods are separated from the original dataset as test-sets. The index series of data are separated from both input and target datasets that are free of any missings (-999) values. This exclusion of a small batch is important for assesing the performance of the models in the next step. This way, it is possible to evaluate the performance of each prediction function by using a dataset that is not seen by the model during training. The separated periods are among the extreme conditions from winter and summer, as well as a third period with high fluctuation in daily profiles.

Second, generating all possible combinations of encountring missing values in the input dataset: this strategy is applied to input samples to provide a benchmark for comparing the accuracy of each candidate method. Accordingly, a series of random missings with a duration of 24 hours and with dissimilar combinations are embedded into the inputs. Six weather stations are included in input features. As a result, to apply all combinatorics of 0 to six, 64 different states are necessary.

The methodology of this paper demands separate training of models for each weather parameter. A unique training set with all modifications, including the selected features, and intended missing and eliminations, is applied to all candidate methods. In all methods, 15% of input data is held out for the validation set.

Mean square error (MSE), Root Means Square Error (RMSE), and Throughput Rate (TR) are statics measures which are considered to evaluate the accuracy of candidate methods (Eq.1-3). The MSE and RMSE in this step refer to the error between outputs of trained model (predictions) and the target of the models (measurements). The throughput rate is the amount of time that each sample is evaluated by training algorithm method. Since training is repeated a few times for all training algorithms, the lowest MSE and throughput rate is selected as the performance of the predictors.

$$MSE = \frac{1}{n}\sum_{1}^{n}(Y_{Measured} - Y_{predicted})^2 \quad (1)$$

$$RMSE = \sqrt{\frac{1}{n}\sum_{1}^{n}(Y_{Measured} - Y_{predicted})^2} \quad (2)$$

$$TR = Training\ Time/number\ of\ samples \quad (3)$$

Where $Y_{Measured}$ refers to actual measurements of weather parameters (temperature/relative humidity); and, $Y_{predicted}$ is the predicted values by trained functions.

**Result and discussions**

After the pre-processing stage, the training-sets are fed into all machine learning methods for training. Trainings are performed on an Intel 6600U CPU @ 2.60 GHz, while utilizing MATLAB's parallel computing toolbox. Although neural networks can be executed on the Graphics Processing Unit (GPU), this study refrains from using the GPU, to keep consistency between the machine learning algorithms. The statistical measures of the trained models for temperature and relative humidity are presented in Table 2 and Table 3, respectively.

*Table 2 - The training performance for temperature*

| Method | MSE (°C) | RMSE (°C) | Throughput (ms) |
|---|---|---|---|
| ET | **0.32** | **0.56** | 0.72 |
| GPR | 0.37 | 0.61 | 229.11 |
| LR | 0.69 | 0.83 | **0.06** |
| NN | 0.42 | 0.64 | 0.36 |
| RT | 0.46 | 0.68 | 0.15 |
| SVR | 0.73 | 0.86 | 51.51 |

It is observed that for both temperature and relative humidity, Ensemble Tree (ET) returns the smallest error, while Linear regression (LR) training concluded faster than the other methods. The LR is well-recognized as a fast training method, at the cost of lower accuracy. This is visible for both parameters in Table 2 and Table 3, as the LR returns the second-worst error. Gaussian Process regression (GPR) has the second-best performance while being the most time-consuming among all training methods. Neural networks, on the other hand, seem to provide a reasonable balance between accuracy and training time.

*Table 3 - The training performance for relative humidity*

| Method | MSE (%) | RMSE (%) | Throughput (ms) |
|---|---|---|---|
| ET | **7.44** | **2.72** | 0.71 |
| GPR | 10.17 | 3.19 | 42.68 |
| LR | 21.68 | 4.65 | **0.07** |
| NN | 8.87 | 2.97 | 0.25 |
| RT | 10.89 | 3.3 | 0.18 |
| SVR | 23.36 | 4.83 | 28.94 |

After training all prediction functions, in this step, the test sets are engaged in the prediction process to evaluate the response of each method facing new sets of data. As mentioned, these portions of data are separated from the original dataset at the beginning. This further ensures the reliability of analysis in this step.

It is important to note that during real-time predictions, reliance on other stations can have its own risks. Namely, one or more stations that are used as an input to the predictor can also be out of order due to maintenance, or provide anomalous values. Such occurrences, although less likely, are not impossible. This is evident from a simple assessment of missing or anomalous data from other stations, as shown in Table 4. Regardless of which station is out of order, the probability of encountering a single missing or anomalous value from another station is 16% and 32% for temperature and relative humidity, respectively. The chances of having more than one station out of order are significantly lower, although still not insignificant.

Therefore, we manipulated the three test-sets based on the combinations of missing data as discussed in stage two. As a result, 64 combinations of inputs are retrieved and for each combination, the RMSE is calculated. In this step, we focus on the worst performance of each method, since we are interested in the least risky performance of the predictors. These indicators are highlighted in Table 5 and Table 6 for temperature and relative humidity, respectively. It is observed that the accuracy of predictors can vary based on the context and conditions, as some models perform better in extreme summer periods and others in extreme winter conditions. However, the neural network displays the best overall performance as shown in Table 5 and Table 6.

**Conclusion**

This study proposes a method for real-time prediction of missing weather data to facilitate online energy simulation. The proposed method (neural network) is contrasted against six other supervised machine learning methods.

Initially, the methods are compared based on performance and training time. Also, a robust investigation of the accuracy of the candidate methods is performed by indexing three extreme periods. It is observed that the performance of all methods strongly depends on the quantity of the missing values in supplementary stations. However, the neural network proves to outperform all other machine learning methods.

## Acknowledgment

This study was initially supported by Fondazione Fratelli Confalonieri with the scholarship "Borsa per dottorandi di ricerca delle Universita Milanesi" funded by Fondazione Fratelli Confalonieri.

*Table 4- Probability of encountering missing data in other weather stations*

|  | 6 missing | 5 missing | 4 missing | 3 missing | 2 missing | 1 missing | No missing |
|---|---|---|---|---|---|---|---|
| Temperature | 0.012 | 0.0025 | 0.0088 | 0.3 | 2.5 | 16.1 | 81 |
| Relative humidity | 0.0076 | 0.0039 | 0.027 | 0.99 | 6.9 | 32.5 | 59.6 |

*Table 5- Comparison between candidate supervised machine learning methods - RMSE Temperature (°C)*

|  | Method | 6 missing | 5 missing | 4 missing | 3 missing | 2 missing | 1 missing | No missing |
|---|---|---|---|---|---|---|---|---|
| **Test-set** | EB | 13.7 | 13.7 | 11.9 | 7.5 | 4.8 | 1 | **0.6** |
|  | GPR | 11.2 | 14.7 | 15.1 | 15 | 11.8 | 0.7 | **0.6** |
|  | LR | **7.1** | 7.1 | 7.1 | 6.9 | 6.5 | 3.7 | 0.7 |
|  | NN | 8 | **6** | **2.5** | **1.8** | **1.4** | **0.9** | **0.6** |
|  | RT | 7.2 | 7.2 | 7.2 | 7.2 | 7.2 | 4.2 | 0.7 |
|  | SVR | 7.6 | 7.8 | 7.7 | 7.5 | 6.9 | 4.7 | 0.7 |

*Table 6- Comparison between candidate supervised machine learning methods - RMSE relative humidity (%)*

|  | Method | 6 missing | 5 missing | 4 missing | 3 missing | 2 missing | 1 missing | No missing |
|---|---|---|---|---|---|---|---|---|
| **Test set** | EB | 9.3 | 10.8 | 7.8 | 5.9 | **5.1** | 4.5 | 4.2 |
|  | GPR | 9.6 | 25.9 | 18.1 | 13.9 | 11.5 | **4.3** | 3.7 |
|  | LR | 10.9 | 11.2 | 10.7 | 10.1 | 9.4 | 7.8 | 3.7 |
|  | NN | **7.6** | **6.9** | **5.7** | **5.3** | 5.6 | **4.3** | **3.6** |
|  | RT | 9.7 | 14.3 | 12.9 | 11.9 | 6 | 5 | 4.2 |
|  | SVR | 11.3 | 11.8 | 11.5 | 11.1 | 10.3 | 8.8 | 3.9 |